\documentclass[a4paper]{article}

\usepackage{INTERSPEECH_v2}
\usepackage{color}
\usepackage{multirow}
\usepackage{slashbox}

\title{Effects of Word Embeddings on Neural Network-based Pitch Accent Detection}
\name{Sabrina Stehwien, Ngoc Thang Vu, Antje Schweitzer}
\address{University of Stuttgart, Germany}
\email{\{sabrina.stehwien,thang.vu,antje.schweitzer\}@ims.uni-stuttgart.de}

\begin{document}

\maketitle
\begin{abstract}
Pitch accent detection often makes use of both acoustic and lexical features based on the fact that pitch accents tend to correlate with certain words.
In this paper, we extend a pitch accent detector that involves a convolutional neural network to include word embeddings, which are state-of-the-art vector representations of words.
We examine the effect these features have on within-corpus and cross-corpus experiments on three English datasets.
The results show that while word embeddings can improve the performance in corpus-dependent experiments, they also have the potential to make generalization to unseen data more challenging.
\end{abstract}

\section{Introduction}

Automatically detecting pitch accents from transcribed speech is a well-established field of research \cite{Wightman1994,Hirschberg1993,Rosenbergphd}.
Most approaches use supervised statistical learning methods on time-aligned data and focus on labelling either words or syllables. Best results are obtained when using good-quality speech corpora that have been manually transcribed and prosodically annotated. The acoustic features are fairly standardized, usually consisting of pitch and energy values extracted per frame or aggregated across segments.
Most methods benefit from the addition of lexical features such as part-of-speech tags \cite{Rosenberg2012,Schweitzer2009,Anantha2008,Sun2002,Chen2004}.
Studies on the relationship between text and prosody in automatic detection methods dates back to early research in the field \cite{Pan1999,Hirschberg1993}.
Research on cross-genre prominence detection has also taken textual features into account \cite{Yuan2005,Margolis2010}.
The prediction of prosodic prominence from text has been widely investigated for text-to-speech (TTS) synthesis; for example, in \cite{Nenkova07} it was shown that certain words are accented quite frequently, and that this has an effect on prediction performance.

In a previous study we have introduced a model that uses a convolutional neural network (CNN) as a simple and efficient way of detecting pitch accents \cite{StehwienVu2017a}. This method has the advantage that very little preprocessing is necessary due to the fact that the input data consists of low-level acoustic features.
In line with the paradigm of using information that is as low-level as possible and letting the model learn feature representations of its own, we test an extension of this model that includes word embeddings \cite{Glove,w2v} as lexical information. 
Word embeddings are vector representations of words that are a standard method of encoding text in state-of-the art neural network models. Trained on large amounts of data in an unsupervised fashion, neural network algorithms ``embed'' words into vector space according to their textual context. Word embeddings have shown to encode both syntactic and semantic similarity \cite{Mikolov2013linguistics}.
While research in neural-network based TTS synthesis has made use of word embeddings to predict prominence from text \cite{Wang04,Rendel2016}, to the best of our knowledge this paper is the first to test word embeddings for pitch accent detection on transcribed speech.
We examine the effect of extending our acoustic model with word embeddings to within-corpus and cross-corpus experiments on three English corpora that are manually annotated with reference pitch accents.
Our results show that adding word embeddings increases pitch accent detection performance on within-corpus experiments. In cross-corpus experiments, it is found that word embeddings are strong features that have the potential to overfit, thus making generalization more challenging.

\section{Datasets}

All datasets used in this work have been previously used for pitch accent detection and have been manually annotated with ToBI labels \cite{Silverman}.
For our experiments, in which we aim to classify words as being pitch accented or not, the ToBI labels are grouped together into two classes: \textit{accented} and \textit{none}. Unsure events, that is events where the annotator was not certain of the presence of a pitch accent, are grouped into the \textit{none} class. 
Table~\ref{tab:corpora} lists the number of words and accented words in each dataset used in our experiments.

The first datasets is a subset of the Boston University Radio News Corpus (BURNC) \cite{Ostendorf1995} that was recorded from 3 female and 2 male speakers. 
The Boston Directions Corpus (BDC) \cite{BDC} contains monologues by 4 speakers (3 male, 1 female) giving directions, both as read and spontaneous speech.
The third corpus is the LeaP corpus \cite{Gut2002}, collected for prosodic research on non-native speech.
In our experiments we use part of the LeaP corpus that consists of read stories and story retellings by non-native speakers with different L1 backgrounds. The subset contains speech recorded from 35 different speakers (22 female, 8 male) across varying proficiency levels. In total the subset comprises 14 different L1 backgrounds. 
Previous work \cite{Levow2009} has shown that despite the lower proficiency, the automatic pitch accent detection performance on LeaP is comparable to that on native speech.

\begin{table}[th]
	\caption{Dataset statistics}
  \label{tab:corpora}
  \centering
  \begin{tabular}{l||l|l|l}
    \toprule
	  corpus  & words & accented & majority class \\
	\hline
	  BURNC & 26742 & 13780 & 51.5\% accented\\
	  BDC & 19225 & 8646 & 55.0\% accented\\
	  LeaP  & 14651 & 6340  & 56.7\% none \\
    \bottomrule
  \end{tabular}  
\end{table}

\section{Baseline Acoustic Model}

The following section describes the supervised learning model that is trained to label each word as carrying a pitch accent or not. A schematic overview of the baseline architecture is shown in the left part of Figure~\ref{fig:model}.

\subsection{CNN-based model}

The baseline model consists of a 2-layer convolutional neural network (CNN), following the model previously introduced for this task \cite{StehwienVu2017a}.
The CNN takes as input frame-based acoustic features representing the current word and its right and left direct context words.
The signal is divided into $s$ overlapping frames and represented by a $d$-dimensional feature vector for each frame. 
Thus, for each word, a matrix $W \in R^{d \times s}$ is formed as input. 
The first convolution layer consists of 100 2-dimensional kernels of the shape $6 \times d$ and a stride of $4 \times 1$, with $d$ as the number of features. The kernels encompass the whole feature set to ensure that all features are learnt simultaneously. The second layer consists of 100 kernels of the shape $4 \times 1$ and a stride of $2 \times 1$.
After applying two convolution layers, max pooling is used to find the most salient features.
The max pooling size is set so that the output of each max pooling on each of the 100 feature maps has the shape $x=100$.
Finally, we apply dropout \cite{Srivastava14} with $p=0.2$, which means that 20\% of the neurons are corrupted in order to avoid overfitting. We also use $l$2 regularization.
The resulting feature representation is fed into the softmax layer that consists of 2 units to form a binary classifier. 

\subsection{Acoustic features}

The acoustic features are extracted from the speech signal using the OpenSMILE toolkit \cite{opensmile}.
The features are 6 low-level acoustic descriptors of pitch, intensity and amplitude as provided in the OpenSMILE default feature sets: root mean square signal frame energy, loudness, smoothed F0, voicing probability of the F0 candidate, harmonics-to-noise ratio and zero-crossing rate.
The intensity (energy, loudness) features are computed for each 20 ms frame and the remaining features for each 50 ms frame. All are computed with a 10ms shift.\footnote{This frame size combination proved best in preliminary experiments.} The features do not require speaker-normalization \cite{StehwienVu2017a}.
The time intervals that indicate the word boundaries provided in the corpus are used to create the input feature matrices by grouping all frames for each word into one matrix.
Afterwards, zero padding is added to the end of the input matrix to ensure that all matrices have the same size.

\begin{figure}
\includegraphics[width=.45\textwidth]{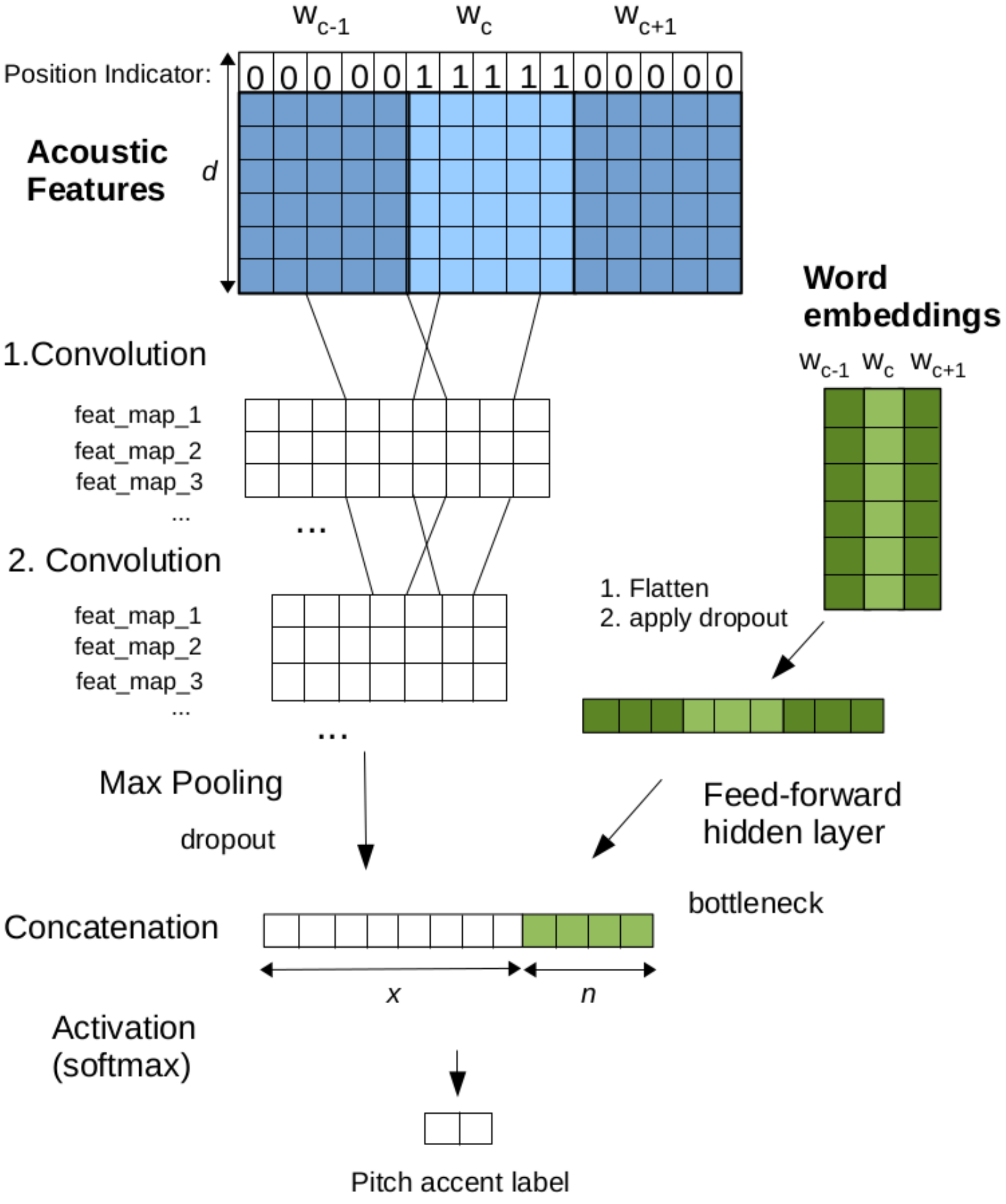}
\caption{Model overview for pitch accent detection using a CNN for acoustic features and extended by adding a feed-forward network for word embeddings, using an input context window of 3 words.}
\label{fig:model}
\end{figure}

Including the context words in the feature matrix introduces information that does not pertain to the current word. For this reason, it was found \cite{StehwienVu2017a} that the model greatly benefits from position features (or indicators) that are appended as an extra feature to the input matrices (see Figure~\ref{fig:model}). These features indicate the parts of the matrix that represent the current word. The rest of the matrix consists of zeros in this dimension. In the first convolution layer we ensure that the kernels always include the position indicator, hereby keeping the model informed which frames belong to the current word.

\section{Lexico-Acoustic Model}

\subsection{Model extension}

The lexico-acoustic model is an extension of the baseline acoustic model. The extension consists of an additional feed-forward network that learns features from word embeddings, shown in Figure~\ref{fig:model} (in green shading).
The input consists of one word embedding vector per input word. The vector values are used as non-trainable matrix weights.
This is fed into the hidden layer, which has only a few output units $n$ that form a ``bottleneck''.
Dropout with $p=0.8$ is applied to the input before feeding it through the hidden layer.
Afterwards, the two output layers of the acoustical and lexical models are regularized separately before they are concatenated into one final feature representation. This is fed into the softmax classification layer.

\subsection{Word embeddings}
\label{sec:oov}

As input features, we test two different types of word embeddings: GloVe \cite{Glove} and word2vec \cite{w2v}.
The respective algorithms are usually applied to very large datasets to train the embeddings. Since the datasets used in this paper are comparatively small, we use the pre-trained 300-dimensional versions available online.
These embeddings are used to represent all words in the datasets. Both GloVe and word2vec yield a number of out-of-vocabulary (OOV) words. We removed noise from the word labels such as special characters and we split contractions, e.g. \textit{she'll} $\rightarrow$ \textit{she} and hyphenated words, e.g. \textit{eighty-eight} $\rightarrow$ \textit{eight}. In both cases we kept the part of the word expected to be relevant for the presence or absence of a pitch accent. This way, the OOV rate was reduced to the numbers shown in Table~\ref{tab:oov}. We set the embedding vectors for OOV words to consist of ones.\footnote{This way we ensure that OOV embeddings receive a separate representation without having to custom-train them.}
The word2vec pre-trained embeddings do not include the frequent words \textit{a, and, of} and \textit{to}, referred to as stopwords. We take this difference in to account in Table~\ref{tab:oov}. These stopwords make up most of the OOV words in the word2vec embeddings, and are very rarely accented. The remaining OOV words, however, are frequently accented. The effect of stopwords represented as OOV in word2vec is discussed in section \ref{sec:stopwords}.

As input information we use unigram and trigram word embeddings, that is we use the word embedding vector of either only the current word or along with its two neighbouring words.
The latter case also requires zero padding if no neighbouring word is present, as is the case at the end of an utterance or speaker turn.
We test different bottleneck sizes $n$, so that the final feature representation along with the acoustic features of size $x=100$ has the dimensionality $100+n$.
The sizes are set such that the output of the lexical model is much smaller than that of the acoustic model, for example $n={10,20,30}$.
By restricting the number of lexical features, we aim to ensure that it does not overpower the acoustic information.

\begin{table}[t]
\caption{Out-of-vocabulary words for both word embedding types on all three datasets.}
  \label{tab:oov}
  \centering
	\begin{tabular}{l||l|l|l}
    \toprule
		 & BURNC & BDC & LeaP \\
	\hline
		GloVe OOV & & & \\
		tokens & 233 & 19 & 4 \\
		types & 64 & 11 & 4 \\
		accent rate & 93\% & 74\% & 50\% \\
		\hline
		w2v OOV &  & & \\
		tokens & 3375 & 2493 & 1822 \\
		types & 231 & 66  & 6 \\
		stopword rate & 86.5\%  & 87\% & 99.9\% \\
		accented stopwords & 3\% & 13\% & 6\% \\
		accented remaining & 81.9\% & 83\% & 100\% \\
	 \bottomrule
  \end{tabular}  
\end{table}

\section{Experiments}

\subsection{Within-corpus and cross-corpus training}

The models are trained usind 10-fold cross-validation for 20 epochs with an adaptive learning rate (Adam \cite{Adam17}). After training, the best model according to a held out validation set is applied to the test set. All experiments are repeated 5 times and the results are averaged.

It is to be noted that the splitting of the corpora into training and test sets are carried out randomly, which means that the within-corpus experiments are neither evaluated independent of speaker nor of lexical content. The latter can be found especially in the BURNC and LeaP corpora, in which different speakers often read the same text. In the LeaP dataset, around half of the utterance files were recorded from speakers reading out the same story. Since the experiments focus on adding lexical information, we do expect the model to learn observed textual features. For this reason, strictly unseen data is tested in the cross-corpus experiments, using entirely different corpora. 

For the within-corpus experiments, we split the datasets into 10 fixed cross-validation splits, and use 1000 held-out words from the respective training split as the development set.
In the cross-corpus experiments, the same 10 test sets are used for comparability with the within-corpus versions. Training is carried out on an entire source corpus and the development set consists of 1000 held-out words from this dataset. In the \textit{ALL} setting, the separate target training set for each target test split is added to the training data.

\subsection{Results}

\subsubsection{Baseline}

Table~\ref{tab:all} shows the accuracy obtained in the within-corpus and cross-corpus experiments.
As expected, we observe that the performance in the cross-corpus settings drops when compared to the within-corpus setting.
BURNC appears to be by far the easiest corpus for pitch accent detection, resulting in accuracy levels obtained while training on BDC or LeaP that are higher than those obtained on the respective within-corpus test sets.
The \textit{ALL} experiments show a slight rise in accuracy that does not quite reach the baseline within-corpus performance, except on LeaP.
The BDC datasets appears to be the most difficult to model using this method.
Overall, the acoustic model can be said to generalize reasonably well across corpora and genres and the within-corpus detection accuracy can be compared to that reported in previous work:
The results on BURNC, outperform previous methods \cite{Rosenbergphd}.
The performance on BDC is slightly higher than the results obtained using acoustic features and speaker-independent evaluation in \cite{Sridhar2008}. The accuracy on the LeaP corpus is similar to previously reported numbers obtained at the syllable level \cite{Levow2009}, although these experiments are not directly comparable.

\begin{table}[t]
	\caption{Results (accuracy in \%) for within-corpus and cross-corpus pitch accent detection using the baseline acoustic model, with added unigram GloVe word embeddings and 10 bottleneck features, and using only the embeddings as features (shown in italics). The results are averaged across 10 cross-validation splits and 5 repetitions of each experiment.}
  \label{tab:all}
  \centering
	\begin{tabular}{l||l|l|l}
		\toprule
		\backslashbox{Train}{Test} & BURNC & BDC & LeaP \\
	\hline
		BURNC & & & \\
		acoustic 	& 87.1 & 74.2 & 79.2 \\
		acoustic+embs 	& \textbf{87.5} & 75.5  & 78.6 \\
		\textit{embs-only} 	& \textit{78.5} & \textit{71.6} & \textit{76.0}  \\
	\hline
		BDC & & & \\
		acoustic	& 82.3 & 78.0 & 76.3 \\
		acoustic+embs	& 82.6 & \textbf{81.2}  & 77.5 \\
		\textit{embs-only} 	& \textit{75.0} & \textit{76.0} & \textit{74.5} \\
	\hline
		LeaP & & & \\
		acoustic	& 82.6 & 72.1 & 80.5 \\
		acoustic+embs 	& 77.7 & 73.0 & \textbf{83.5} \\
		\textit{embs-only} 	& \textit{67.7} & \textit{68.0} & \textit{80.9} \\
	\hline
		ALL & & & \\
		acoustic 	& 86.6 & 77.4 & 80.8 \\
		acoustic+embs 	& 87.0 & 80.6 & 83.4 \\
		\textit{embs-only} 	& \textit{75.2} & \textit{72.7} & \textit{77.6} \\
	 \bottomrule
  \end{tabular}  
\end{table}

\subsubsection{Effect of adding word embeddings}

The first experiments compare the effect of adding word embeddings to within-corpus and cross-corpus settings. We use unigram word embeddings and 10 bottleneck features because these are taken to be less corpus-specific than trigram features. Table~\ref{tab:all} contains only the results using GloVe embeddings since the results using word2vec were similar.
We observe an increase in accuracy on all corpora (marked in bold) in the corpus-dependent setting, which shows that the model can learn representations from word embeddings that are helpful for detecting pitch accents.
In contrast, the results obtained on cross-corpus experiments lead to the conclusion that adding word embeddings to the baseline model cannot reliably improve performance. The results obtained from training on the LeaP corpus even show a large drop in accuracy when using word embeddings. This is due to the strong lexical features causing the model to overfit to the respective training corpus. A larger improvement on BDC and LeaP can can be observed when target corpus data held out from the test set is seen in training (\textit{ALL}). In fact, the performance of \textit{ALL} on LeaP is as high as the within-corpus results using word embeddings. The results on BURNC, however, are largely unaffected by these changes. 
The embeddings-only results were added to give an impression of how these features would perform alone. In all cases, these features alone are not as strong as the acoustic features alone, except on the LeaP corpus. This confirms that the lexical overlap in LeaP greatly facilitates the detection of pitch accents.

\subsubsection{Effect of N-gram and bottleneck sizes} 

In Table~\ref{tab:emb-within} we show results when adding unigram and trigram embeddings to the baseline acoustic model. We also test different bottleneck sizes $n$.
The word embeddings improve the accuracy over the baseline on all corpora across all tested settings.
Overall, the results do not show a specific pattern, from which it can be concluded that neither the choice of N-gram and bottleneck size, nor the embedding type is critical in our experiments.
The corpora that benefit the most from these lexical features are LeaP and BDC. Here, we observe an increase in around 3 percentage points in all experiments. We assume that this effect is due to the lexical overlap in LeaP and the lower baseline accuracy in BDC.
To provide more insight into these numbers, we also show the precision, recall and F1 scores for the accent class in Table~\ref{tab:f1} when using unigram embeddings and 10 bottleneck features. The word embeddings lead to a gain in F1, precision and recall in all cases, except for precision on BURNC. 
\begin{table}[t]
\caption{Results (accuracy in \%) for within-corpus pitch accent detection with word embeddings using unigrams, trigrams and varying bottleneck sizes.}
  \label{tab:emb-within}
  \centering
	\begin{tabular}{l||lr|lr|lr}
    \toprule
		Corpus		& \multicolumn{2}{|c|}{BURNC} & \multicolumn{2}{|c|}{BDC} & \multicolumn{2}{|c}{LeaP} \\
		Embeddings      & glove & w2v & glove & w2v & glove & w2v \\
	\hline
		unigram		& & & & & & \\
		$n = 10$	& 87.5 & 87.6 & 81.2 & 80.6 & 83.5 & 83.9 \\
		$n = 20$	& 87.4 & 87.7 & 81.5 & 81.1 & 83.6 & 83.8 \\
	\hline
		trigram	& & & & & & \\
		$n = 10$	& 87.7 & 87.7 & 82.4 & 81.1 & 83.9 & 83.6 \\
		$n = 30$	& 87.8 & 87.5 & 82.7 & 81.4 & 83.7 & 83.8 \\
    \bottomrule
  \end{tabular}  
\end{table}

 \begin{table}[t]
\caption{Precision, recall and F1 scores for the accent class using unigram word embeddings and 10 bottleneck features.}
  \label{tab:f1}
  \centering
	\begin{tabular}{l||l|l|l}
    \toprule
		Experiment & Precision & Recall & F1 \\
	\hline
		BURNC & & \\
		baseline & 88.0 & 86.7  & 87.3 \\
		w2v 	 & 87.2 & 88.9  & 88.1 \\
		GloVe 	 & 88.1 & 87.7  & 87.7 \\
		\hline
		BDC & & \\
		baseline & 74.5 & 77.8  & 76.0  \\
		w2v 	 & 77.6 & 80.9  & 79.2  \\
		GloVe 	 & 77.4 & 82.5  & 79.8 \\
		\hline
		LEAP & & \\
		baseline & 76.2 & 80.4 & 78.1 \\
		w2v 	 & 79.7 & 84.6 & 82.0  \\
		GloVe 	 & 79.8 & 82.8 & 81.2 \\
	 \bottomrule
  \end{tabular}  
\end{table}

\subsubsection{Performance on stopwords}
\label{sec:stopwords}

In word2vec, stopwords are represented as OOV with a zero vector, wheras in GloVe, they are part of the vocabulary.
Because of this difference we also compute the performance accuracy just on stopwords (Table~\ref{tab:stopwords}).
On the LeaP and BDC datasets, we observe better performance on these words with both types of embeddings.
It appears that stopwords are more likely to be labelled correctly (which is usually not accented) when there is some information about the word identity available, regardless of how they are represented as vectors. 
The results on BURNC do not show a clear pattern. Overall, these numbers appear to be reflective of the overall increase in performance obtained by adding the lexical information.
\begin{table}[t]
	\caption{Stopword accuracy (\%) using unigram embeddings and 10 bottleneck features.}
  \label{tab:stopwords}
  \centering
	\begin{tabular}{l||l|l|l}
    \toprule
		 & BURNC & BDC & LeaP \\
		 \hline
		baseline & 98.2 & 88.9 & 86.9 \\
		GloVe & 98.2 & 92.7 & 94.2 \\
		w2v & 97.8 & 92.7 & 94.3  \\
	 \bottomrule
  \end{tabular}  
\end{table}

\section{Discussion}

The baseline model alone appears to be reasonably strong, despite the fact that the input features require very little preprocessing and linguistic knowledge. Further analysis and optimization steps may increase performance on the BDC dataset. 
Within-corpus experiments show that simply adding word embeddings does provide information that the model can use to learn useful features that correlate with pitch accents, as the gain in accuracy in these cases demonstrates. 
We find, however, that the challenge lies in that such strong features tend to lead to overfitting.
The fact that the more textual overlap is found in training and test data, the better the increase in performance, must be kept in mind when using word embeddings in such a setting.
Possible future directions include the exploration of how the acoustic and lexical information interact during training, and what lexical information the model is learning to exploit: e.g. to what extent semantic or syntactic information is encoded in the bottleneck features.

\section{Conclusion}

In this paper, we investigate the effect of adding word embeddings as a simple method of exploiting lexical information in an extension of a CNN-based pitch accent detector.
The results from within-corpus experiments show that the model can clearly benefit from word embeddings, but that generalizing to unseen datasets remains a challenging task. 
Cross-corpus experiments show that the baseline acoustic model alone yields good results. Future research can help ascertain what type of information the model is learning from word embeddings.

\section{Acknowledgements}
This work is funded by the Sonderforschungsbereich (collaborative research center) SFB-732 of the German National Science Foundation (DFG). 


\bibliographystyle{IEEEtran.bst}

\bibliography{paslu}

\end{document}